\documentclass{article}

\usepackage{PRIMEarxiv}

\usepackage[utf8]{inputenc} 
\usepackage[T1]{fontenc}    
\usepackage{hyperref}       
\usepackage{url}            
\usepackage{booktabs}       
\usepackage{amsfonts}       
\usepackage{nicefrac}       
\usepackage{microtype}      
\usepackage{lipsum}
\usepackage{fancyhdr}       
\usepackage{graphicx}       
\usepackage{float}
\graphicspath{{media/}}     

\title{Human Motion Predicition, Reconstruction, and Generation}

\author{Canxuan Gang~~~~Yiran Wang\\\newline\\
AI Geeks\\\newline\\
\url{https://aigeeksgroup.github.io}
}

\begin{document}
\maketitle

\begin{abstract}

This report reviews recent advancements in human motion prediction, reconstruction, and generation. Human motion prediction focuses on forecasting future poses and movements from historical data, addressing challenges like nonlinear dynamics, occlusions, and motion style variations. Reconstruction aims to recover accurate 3D human body movements from visual inputs, often leveraging transformer-based architectures, diffusion models, and physical consistency losses to handle noise and complex poses. Motion generation synthesizes realistic and diverse motions from action labels, textual descriptions, or environmental constraints, with applications in robotics, gaming, and virtual avatars. Additionally, text-to-motion generation and human-object interaction modeling have gained attention, enabling fine-grained and context-aware motion synthesis for augmented reality and robotics. This review highlights key methodologies, datasets, challenges, and future research directions driving progress in these fields.

\end{abstract}


\section{Human Motion Prediction}

Human Motion Prediction is a fundamental task in computer vision and computer graphics that aims to forecast future human poses and movements based on historical motion sequences. It plays a crucial role in applications such as animation, robotics, autonomous driving, and human-computer interaction. The task involves learning temporal patterns and dependencies in human motion, often using deep learning models like recurrent neural networks (RNNs), transformers, or diffusion models.

A key challenge in human motion prediction is modeling complex, nonlinear motion dynamics while maintaining realism and temporal consistency. Factors such as occlusions, variations in motion styles, and external forces further complicate prediction. Recent advancements leverage attention mechanisms and generative models to improve accuracy and diversity in motion forecasting. By capturing fine-grained spatial-temporal dependencies, modern approaches enhance applications ranging from virtual avatars to predictive healthcare. Ultimately, robust human motion prediction enables more natural and intelligent interactions between humans and AI systems.

\subsection{Generating Smooth Pose Sequences for Diverse Human Motion Prediction (ICCV 2021)}

Predicting diverse and realistic human motion is crucial for applications in animation, robotics, and autonomous systems \cite{mao2021generating}. This paper introduces a unified generative framework that addresses both \textbf{diverse motion prediction} and \textbf{controllable motion generation} by leveraging a pose prior and a structured generative process. 

The authors propose a sequential generation approach where different body parts are predicted in a structured order, allowing the model to maintain smoothness and realism. Unlike previous works that use multiple mappings for diversity, this method relies on a normalizing flow-based pose prior and a joint angle loss to ensure motion feasibility. The approach also introduces a diversity-promoting loss to encourage a wide range of possible motions while preserving physical validity.

Experiments on standard datasets, Human3.6M \cite{ionescu2013human3} and HumanEva-I, demonstrate that the proposed model outperforms state-of-the-art baselines in both accuracy and diversity. The key findings show that generating body parts sequentially leads to more realistic human motion sequences, and smaller patches of motion data can be used effectively with reduced computational overhead.

\begin{figure}[h]
    \centering
    \includegraphics[width=0.8\textwidth]{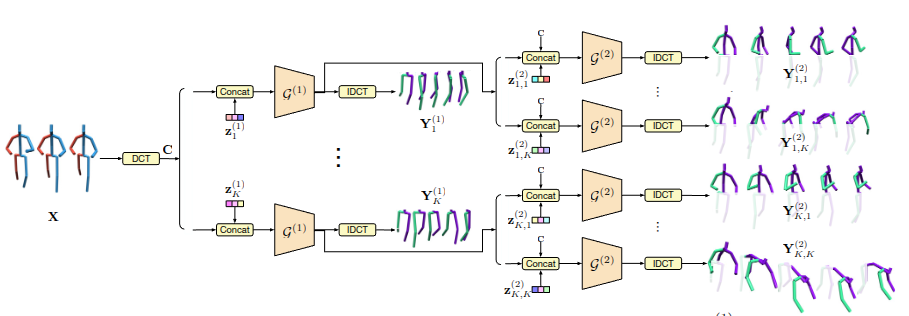}
    \caption{Overview of the proposed method: The model generates diverse and smooth motion predictions by leveraging a normalizing flow-based pose prior and a structured generation process.}
    \label{fig:method}
\end{figure}

This framework significantly advances human motion prediction by providing a unified, controllable, and diverse generative approach that enhances both realism and computational efficiency.

\subsection{Human Motion Prediction under Unexpected Perturbation (CVPR 2024)}

This paper introduces a novel approach to human motion prediction under unexpected physical perturbation \cite{yue2024human}. Unlike traditional motion prediction, which deals with planned or controlled movements, this task focuses on reactive movements caused by unanticipated external forces, such as a push. The study addresses challenges such as data scarcity and the complexity of predicting motion propagation across individuals in a group. To solve these issues, the authors propose the \textbf{Latent Differentiable Physics (LDP)} model, which combines differentiable physics with deep neural networks. The core of the model is the \textbf{Inverted Pendulum Model (IPM)} \cite{gage2004kinematic}, which governs the balance recovery of individuals in response to perturbations. This model is highly data-efficient, requiring minimal data to make accurate predictions.

The LDP model is capable of predicting both individual motions and complex multi-person interactions. Experiments show that LDP outperforms existing baselines in terms of prediction accuracy and generalization, with improvements of up to 70\%. Additionally, the model provides strong interpretability, offering plausible explanations for the predicted motions based on the physical forces involved. The proposed method has significant potential for applications in biomechanics, crowd simulation, and human-robot interaction.

\begin{figure}[h]
\centering
\includegraphics[width=0.8\textwidth]{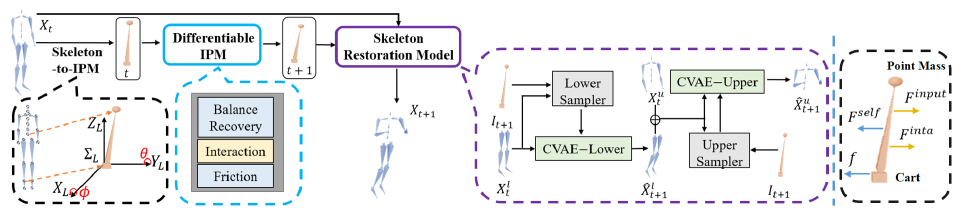}
\caption{Overview of the Latent Differentiable Physics (LDP) model. The model maps full-body poses to the Inverted Pendulum Model (IPM) state, simulates the interaction forces, and reconstructs the full-body motion, enabling effective prediction under physical perturbations.}
\label{fig:motion_prediction}
\end{figure}

\subsection{Harmonizing Stochasticity and Determinism: Scene-responsive Diverse Human Motion Prediction (NIPS 2024)}

This paper introduces \textbf{DiMoP3D}, a novel framework for diverse human motion prediction in 3D scenes \cite{louharmonizing}. Traditional methods of human motion prediction have primarily focused on generating deterministic sequences or predicting multiple motions without accounting for real-world environmental constraints. These methods often fail to integrate the dynamic constraints imposed by physical scenes, such as avoiding obstacles or interacting with objects realistically. DiMoP3D addresses this gap by harmonizing the stochastic nature of human motion with the deterministic constraints of surrounding 3D environments.

DiMoP3D consists of a three-component system: the \textbf{Context-aware Intermodal Interpreter}, which analyzes the scene and detects potential interaction targets; the \textbf{Behaviorally-consistent Stochastic Planner}, which generates obstacle-free trajectories and end-pose predictions; and the \textbf{Self-prompted Motion Generator}, which uses a denoising diffusion model to generate diverse, physically consistent motions. This architecture ensures that predicted motions are both diverse and realistic, conforming to the 3D environment.

The proposed framework is evaluated on two real-world datasets, \textbf{GIMO} \cite{zheng2022gimo} and \textbf{CIRCLE} \cite{araujo2023circle}, showing significant improvements over existing methods in both physical consistency and prediction accuracy. By integrating human intentions and the surrounding environment, DiMoP3D outperforms state-of-the-art methods, particularly in generating motion predictions that align with realistic human-object interactions.

\begin{figure}[h]
\centering
\includegraphics[width=0.8\textwidth]{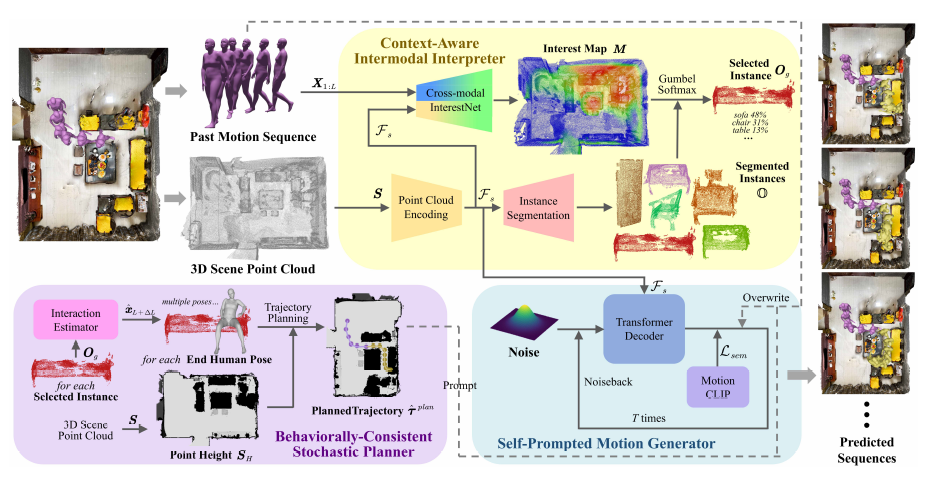}
\caption{Overview of the DiMoP3D architecture. The system incorporates a Context-aware Intermodal Interpreter to analyze 3D scenes, a Behaviorally-consistent Stochastic Planner for planning motion trajectories, and a Self-prompted Motion Generator to produce diverse and physically consistent motion sequences.}
\label{fig:dimop3d}
\end{figure}

\subsection{Contact-aware Human Motion Forecasting (NIPS 2022)}

This paper addresses the challenge of scene-aware 3D human motion forecasting, which aims to predict future human poses while accounting for human-scene interactions \cite{mao2022contact}. Previous methods have struggled to integrate scene context effectively, often leading to artifacts such as “ghost motion.” The proposed approach explicitly models human-scene contacts through per-joint contact maps. These maps capture the distances between human joints and scene points at each time step, providing detailed information about human-scene interactions.

The method consists of a two-stage pipeline. The first stage predicts future contact maps based on past contact maps, past human motion, and the 3D scene. The second stage forecasts the future human poses by conditioning them on the predicted contact maps. The approach ensures consistency between global motion and local poses, which helps avoid unrealistic motion artifacts. Experiments on synthetic and real datasets show that the proposed method outperforms state-of-the-art human motion forecasting models, yielding more accurate predictions for both global translations and local poses.

The framework demonstrates significant improvements over existing methods, particularly in terms of capturing fine-grained human-scene interactions and ensuring temporal consistency in predicted motions. This work highlights the importance of explicitly modeling human-scene contact relationships for more realistic and accurate motion forecasting.

\begin{figure}[h]
    \centering
    \includegraphics[width=0.8\textwidth]{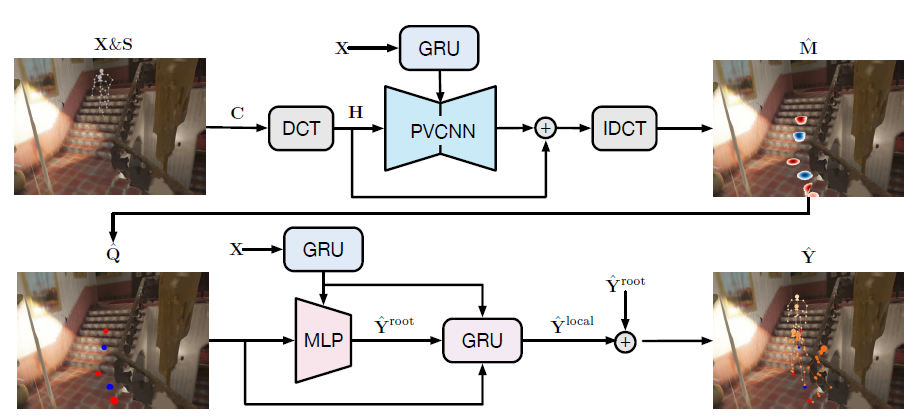}
    \caption{Overview of the contact-aware human motion forecasting approach. The system predicts future contact maps (left) and uses these maps to forecast future human poses (right), ensuring consistency between the global motion and local poses.}
    \label{fig:motion_forecasting}
\end{figure}

\section{Human Motion Reconstruction and Tracking}

Human Motion Reconstruction and Tracking is a fundamental task in both computer vision and computer graphics, focusing on capturing and reconstructing human body movements from visual data. The goal is to generate a 3D representation of a human’s posture and movement over time, using either monocular or multi-view cameras, depth sensors, or other tracking devices. 

Motion reconstruction involves estimating the 3D pose of a human from 2D projections or sparse data, often requiring algorithms that infer the shape and position of the body in space. Tracking refers to continuously estimating the motion of a subject in video or sensor data, ensuring temporal consistency across frames. 

This task has significant applications in fields such as animation, virtual reality, augmented reality, healthcare, and human-computer interaction. Advancements in deep learning and AI have substantially improved the accuracy and efficiency of motion reconstruction and tracking, enabling more realistic simulations and practical real-time systems.

\subsection{Humans in 4D:
Reconstructing and Tracking Humans with Transformers (ICCV 2023)}

This paper introduces a novel approach for reconstructing and tracking humans in video, referred to as \textbf{4DHumans} \cite{goel2023humans}. At its core, the method leverages a fully transformer-based architecture, HMR 2.0, for 3D human mesh recovery from single images. HMR 2.0 advances the state-of-the-art in 3D body pose and shape reconstruction, particularly for challenging and unusual poses. Previous models often struggle with difficult poses or occlusions, but HMR 2.0 provides robust reconstructions across a variety of viewpoints.

To track people over time, the paper presents a 3D tracking system that builds on the HMR 2.0 reconstructions. This system enables the tracking of multiple people in monocular videos, even when occlusions or detection failures occur. The resulting system, \textbf{4DHumans}, is evaluated on the PoseTrack dataset, demonstrating state-of-the-art performance in human tracking. The paper also applies HMR 2.0 to action recognition tasks, where it significantly improves accuracy compared to previous pose-based action recognition methods.

4DHumans combines accurate 3D pose estimation from HMR 2.0 with a robust tracking pipeline that operates across frames, even under challenging conditions. The results show that 4DHumans can accurately track individuals through occlusions and interactions with others, making it a powerful tool for video analysis, robotics, and biomechanics applications. This work demonstrates the potential of transformer-based models to enhance both reconstruction and tracking performance in human motion analysis.

\begin{figure}[h]
\centering
\includegraphics[width=0.8\textwidth]{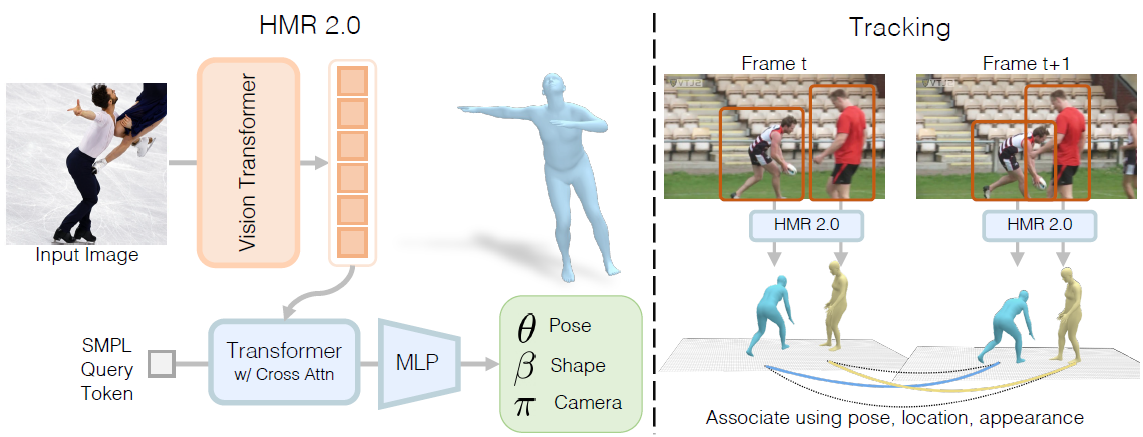}
\caption{Overview of the 4DHumans system. Left: HMR 2.0 for human mesh recovery; Right: 4DHumans for joint reconstruction and tracking in video.}
\label{fig:4DHumans}
\end{figure}

\subsection{RoHM
Robust Human Motion Reconstruction via Diffusion (CVPR 2024)}

This paper introduces RoHM, a novel approach for robust 3D human motion reconstruction from monocular RGB(-D) videos, particularly in the presence of noise and occlusions \cite{zhang2024rohm}. Traditional methods rely on either deep learning models or optimization techniques to reconstruct 3D motions, but these often struggle with global coherence or fail under occlusion. To address these challenges, RoHM utilizes diffusion models, which have demonstrated strong performance in denoising and infilling tasks.

RoHM decouples the problem of motion reconstruction into two sub-tasks: global trajectory reconstruction and local motion prediction. The global trajectory model, TrajNet, refines the root motion, while the PoseNet model focuses on local body motion. These two models are conditioned on noisy inputs, and their outputs are refined using an iterative process to improve accuracy. Additionally, the system integrates a flexible conditioning module, TrajControl, which captures the dependencies between global and local motion spaces to ensure physically plausible motion reconstructions.

The method is shown to outperform state-of-the-art optimization-based techniques, such as HuMoR and MDM, in terms of both accuracy and physical plausibility, especially in scenarios with occluded or noisy input data. The system’s ability to recover realistic 3D human motion in challenging environments is validated on several benchmark datasets, including AMASS and PROX, where it demonstrates superior performance in foot contact prediction and reduces foot skating by a significant margin.

\begin{figure}[h]
\centering
\includegraphics[width=0.8\textwidth]{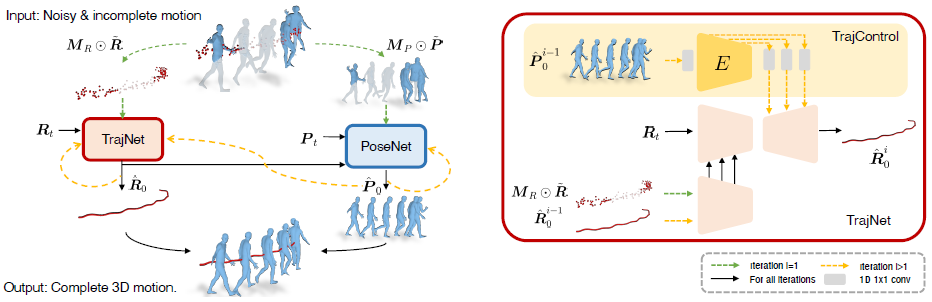}
\caption{Overview of RoHM’s approach. Left: TrajNet for global trajectory denoising; Right: PoseNet for local body motion recovery. The iterative inference process refines both global and local motion.}
\label{fig:rohm}
\end{figure}

\subsection{DiffusionPoser: Real-time Human Motion Reconstruction From Arbitrary Sparse Sensors Using Autoregressive Diffusion (CVPR 2024)}

This paper introduces \textbf{DiffusionPoser}, a real-time human motion reconstruction system capable of utilizing arbitrary combinations of sparse sensors, such as inertial measurement units (IMUs) and pressure insoles \cite{van2024diffusionposer}. Unlike traditional motion capture systems, which rely on a fixed set of sensors, DiffusionPoser offers the flexibility to optimize sensor configurations tailored to the specific activity being monitored, without requiring retraining of the model. The system employs a generative diffusion model for motion reconstruction, which produces realistic motion sequences even from incomplete or noisy sensor data.

DiffusionPoser operates in real-time and utilizes an autoregressive inference scheme to reconstruct human motion. The model is robust to sensor signal corruption or loss, making it ideal for practical applications in health, rehabilitation, and performance tracking. It also supports multiple sensor configurations, optimizing the placement of sensors based on the body regions most relevant to the task, such as pelvis, wrists, or thighs. This adaptability allows users to prioritize sensor comfort and accuracy for specific motions, such as whole-body movements or leg kinematics.

The paper demonstrates the effectiveness of DiffusionPoser through experiments on real-world datasets, showing that it outperforms previous models that rely on fixed sensor configurations. It achieves high accuracy in reconstructing human motion, even when using minimal sensor data, and runs at 20Hz, making it suitable for real-time applications. Additionally, DiffusionPoser can be applied to both the SMPL body model and the OpenSim musculoskeletal model, making it versatile for both general human motion tracking and specialized biomedical research.

\begin{figure}[h]
\centering
\includegraphics[width=0.55\textwidth]{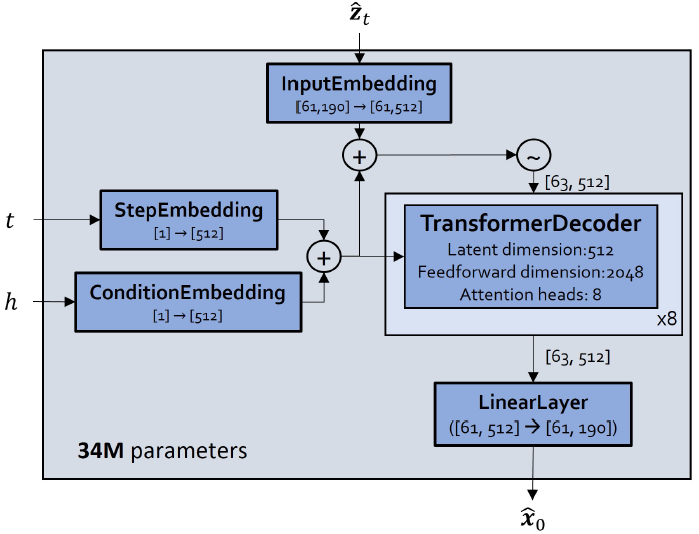}
\caption{Overview of the DiffusionPoser system. It utilizes IMUs and pressure insoles for flexible motion reconstruction, adapting to different sensor configurations without retraining.}
\label{fig:diffusionposer}
\end{figure}

\subsection{Efficient Human Motion Reconstruction from Monocular Videos with Physical Consistency Loss (SA 2023)}

This paper presents a new method for efficiently reconstructing 3D human motion from monocular videos, addressing issues such as foot sliding and jittering that occur in traditional vision-only motion reconstruction \cite{cong2023efficient}. The proposed approach integrates physical constraints into the reconstruction process through a differentiable physical consistency loss, enabling the system to handle complex, highly dynamic movements without relying on computationally expensive bilevel optimization.

The system reformulates human motion dynamics into a single-level trajectory optimization problem, which reduces computational complexity and improves runtime. By introducing a hybrid dynamics formulation, the method optimizes both joint forces and contact forces while maintaining physical realism. The model also optimizes for camera pose uncertainty, compensating for errors in ground plane location, which is crucial for accurate physical inference.

Experiments demonstrate that the proposed method outperforms existing approaches in both speed and accuracy. The method is able to reconstruct complex human motions, such as acrobatic and dynamic movements, in real time from monocular videos, achieving highly plausible physical reconstructions. In addition, an ablation study shows that the physical consistency loss significantly reduces foot sliding and improves motion smoothness.

The key contributions of this method include transforming motion reconstruction into a single-level optimization problem, including joint forces and torques to reduce impacts, and integrating camera pose optimization to enhance reconstruction accuracy. The approach's efficiency is highlighted by its ability to achieve fast convergence even with complex motions, making it a promising tool for real-time human motion reconstruction.

\begin{figure}[h]
\centering
\includegraphics[width=0.8\textwidth]{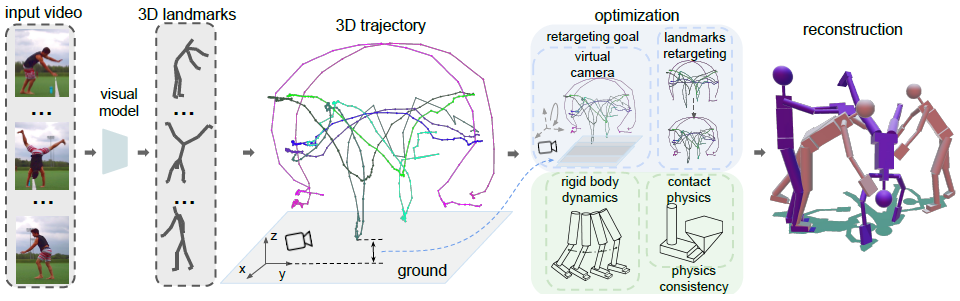}
\caption{Overview of the motion reconstruction pipeline. The system integrates physical consistency loss and optimizes for camera pose uncertainty to reconstruct realistic 3D motion.}
\label{fig:motion_reconstruction}
\end{figure}

\section{Action to Human Motion Generation}

Action to Human Motion Generation refers to the process of synthesizing realistic human motion based on high-level action descriptors or input actions. This task is fundamental in both computer vision and computer graphics as it enables the creation of dynamic human movements for applications in animation, gaming, virtual reality, and robotics.

In this task, an action, such as "walking," "running," or "jumping," is used as input to generate the corresponding motion sequence. The challenge lies in capturing the intricacies of human motion, including body dynamics, spatial constraints, and temporal consistency. Methods often leverage machine learning, particularly deep learning models, to generate realistic motions by learning from large datasets of human actions. These models map abstract action labels or text descriptions to actual body movements, ensuring natural and physically plausible results. The task is critical for creating lifelike virtual characters and for applications requiring human motion synthesis in diverse scenarios.

\subsection{Action2Motion: Conditioned Generation of 3D Human Motions (MM2020)}

This paper introduces Action2Motion, a framework for generating 3D human motion sequences conditioned on action categories \cite{guo2020action2motion}. The task of action-conditioned human motion generation aims to produce diverse and natural 3D motion sequences from predefined action types such as "walking," "lifting a dumbbell," or "throwing." Unlike previous works that rely on initial poses or 2D representations, Action2Motion directly generates 3D motion sequences without prior pose conditions, enabling more realistic and diverse motion synthesis.

The method utilizes a conditional Temporal Variational Auto-Encoder (VAE) with Lie algebra representation for human motion modeling. The Lie algebra representation effectively decouples the anatomy, motion trajectories, and scale of human skeletons, improving motion quality and learning efficiency. The framework uses a posterior network, a prior network, and a generator, enabling the generation of realistic and diverse human motions. The use of a temporal VAE ensures that the generated motions respect temporal dependencies, leading to smoother and more consistent sequences.

The empirical results demonstrate that Action2Motion outperforms existing methods in generating diverse and realistic motions. It achieves high accuracy in recognizing generated motions' action categories and shows significant improvements in motion diversity, as demonstrated on several datasets, including HumanAct12, NTU-RGB-D, and CMU MoCap. The paper also highlights the importance of using Lie algebra representations to generate human motions and provides extensive qualitative and quantitative evaluations of the generated motions.

\begin{figure}[h]
\centering
\includegraphics[width=0.8\textwidth]{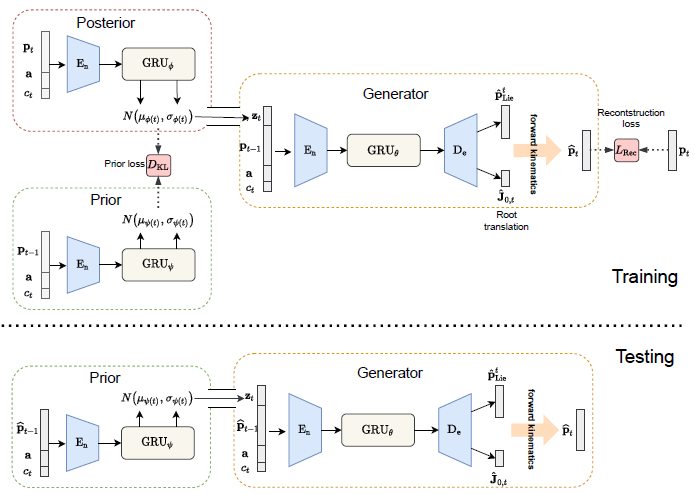}
\caption{Overview of the Action2Motion framework. It utilizes a conditional temporal VAE with Lie algebra representation to generate realistic 3D human motions conditioned on action categories.}
\label{fig:action2motion}
\end{figure}

\subsection{Action-conditioned On-demand Motion Generation (MM2022)}

This paper introduces a framework called \textbf{On-Demand Motion Generation (ODMO)} for generating diverse and realistic 3D human motion sequences conditioned on action types \cite{lu2022action}. Unlike traditional motion generation methods, ODMO is designed to work with just the action category as input, allowing for the generation of long-term motion sequences without requiring initial motion data. This on-demand capability opens new possibilities for customizable motion synthesis based on predefined action categories such as "walking," "jumping," or "dancing."

The framework consists of two main innovations: first, the use of contrastive learning to organize latent space representations of motion sequences, enabling efficient mode discovery within action categories; second, a hierarchical decoding strategy that first reconstructs the motion trajectory and then generates the full motion sequence. The use of a Gaussian Mixture Model (GMM) further enhances the model's ability to generate diverse motions within each action category, making it capable of producing different motion styles on demand.

The model is evaluated on several public datasets (HumanAct12, UESTC, and MoCap) and shows significant improvements over existing methods in terms of both diversity and accuracy. Key features include the ability to perform interpolation between motion modes, customize the motion trajectory via endpoint specification, and generate motions that are both realistic and diverse within a given action category. The system’s flexibility makes it a powerful tool for applications in animation, robotics, and interactive environments.

\begin{figure}[h]
\centering
\includegraphics[width=0.8\textwidth]{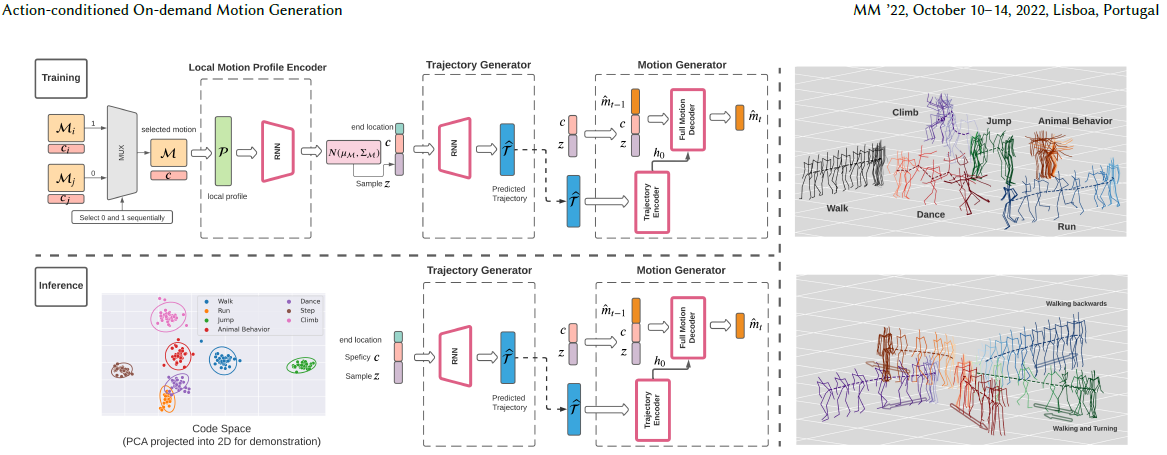}
\caption{Overview of the On-Demand Motion Generation (ODMO) system. The system generates diverse 3D human motion sequences conditioned on action categories, with the capability of trajectory customization.}
\label{fig:odmo}
\end{figure}

\subsection{ActFormer: A GAN-based Transformer towards General Action-Conditioned
3D Human Motion Generation (ICCV 2023)}

This paper introduces \textbf{ActFormer}, a GAN-based Transformer framework for general action-conditioned 3D human motion generation \cite{xu2023actformer}. ActFormer is designed to handle both single-person and multi-person interactive motions, addressing the limitations of previous methods that focus primarily on single-person motions or do not generalize well across different motion representations. The framework leverages the powerful spatio-temporal modeling capabilities of the Transformer architecture and the generative advantages of GANs, combined with a Gaussian Process (GP) latent prior to enforce temporal correlations in generated sequences.

The model uses a two-step process: first, it generates single-person motion sequences conditioned on action labels, and then it extends to multi-person interactive motions by synchronizing the latent vectors shared across persons while differentiating them using positional encodings. To facilitate research in multi-person motion generation, the paper introduces a new dataset of complex multi-person combat actions, complementing existing datasets such as NTU-120 and BABEL.

Experiments show that ActFormer outperforms existing methods in both single-person and multi-person tasks, achieving superior performance on datasets that include a wide range of action categories and interaction complexities. Additionally, the system demonstrates significant improvements in handling multi-person interactions, which are critical for real-world human motion generation applications in areas like animation, robotics, and video synthesis.

\begin{figure}[h]
\centering
\includegraphics[width=0.8\textwidth]{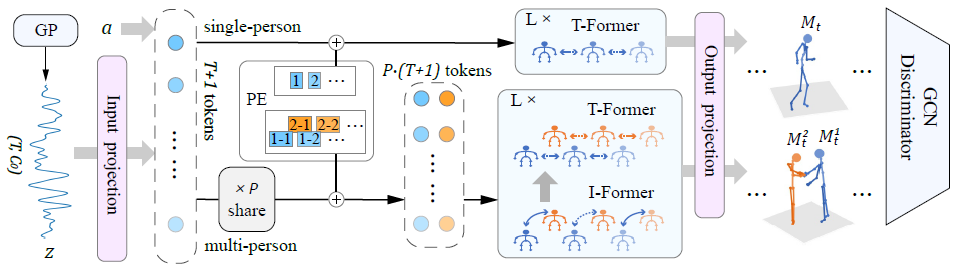}
\caption{Overview of the ActFormer framework. The model generates action-conditioned 3D human motions for both single-person and multi-person settings, leveraging GANs and Transformers.}
\label{fig:actformer}
\end{figure}

\subsection{Action-conditioned 3d human motion synthesis with transformer vae (ICCV 2021)}

This paper presents a novel approach for generating action-conditioned 3D human motion sequences using a Transformer-based Variational Autoencoder (VAE) \cite{petrovich2021action}. The goal is to generate realistic human motions given a semantic action label, such as "throw" or "jump," without requiring initial motion data or poses. The model, named ACTOR, is based on a Transformer architecture that encodes and decodes human body pose sequences represented by the SMPL body model. By training with action-labeled 3D motion data, the model learns to synthesize variable-length motion sequences conditioned on the action type.

The framework works by generating a sequence-level latent representation for motion, which is more efficient than prior methods that operate on frame-level latent embeddings. The model is capable of generating diverse and realistic motions even for actions with limited training data, a significant advantage over previous methods reliant on large motion capture datasets. The paper introduces a new method of using positional encodings for the Transformer decoder to generate sequences of varying lengths without the typical issue of pose regression.

ACTOR is evaluated on several public datasets, including NTU RGB+D, HumanAct12, and UESTC, demonstrating state-of-the-art performance in terms of both diversity and realism of generated motions. The model’s ability to handle monocular motion estimates is particularly notable, as it can generate high-quality motions even from noisy input data. Additionally, the model is shown to be useful in augmenting action recognition tasks and in denoising motion data.

\begin{figure}[h]
\centering
\includegraphics[width=0.8\textwidth]{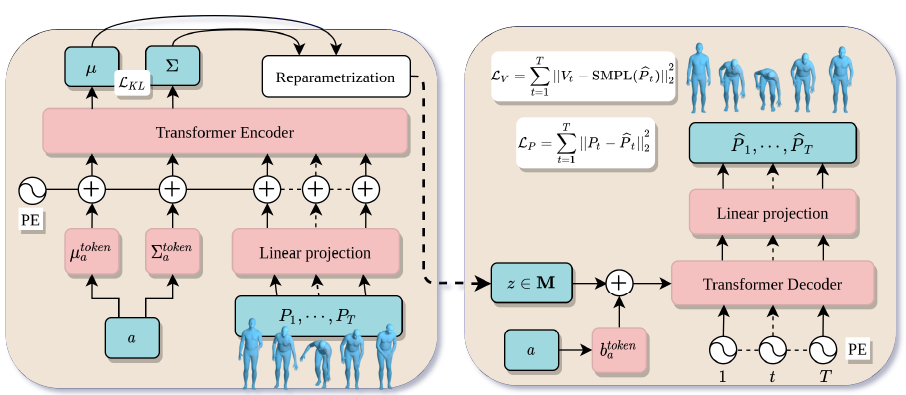}
\caption{Overview of the ACTOR framework. The model generates action-conditioned 3D motion sequences using a Transformer-based VAE architecture.}
\label{fig:actor}
\end{figure}

\section{Text to Human Motion Generation}

Text-to-Human Motion Generation is a fundamental task in computer vision and computer graphics that aims to generate realistic human motion sequences from natural language descriptions. Given a textual prompt—such as \textit{“a person walking slowly with a limp”} or \textit{“a dancer performing a spin”}—the goal is to synthesize a corresponding motion sequence, typically represented as a series of 3D human poses over time.

This task is crucial for applications in virtual avatars \cite{zhang2024motionavatar}, animation, gaming, and human-computer interaction. It involves challenges like understanding complex linguistic instructions, ensuring temporal coherence in generated motions, and maintaining realism in human movement. Recent advancements leverage deep learning models, including Transformers and diffusion-based generative models, to map text inputs to plausible and diverse motion outputs. High-quality datasets and multimodal learning techniques further enhance model performance. By bridging natural language and human motion, this task plays a key role in advancing AI-driven animation and digital human synthesis.

\subsection{Executing your Commands via Motion Diffusion in Latent Space (CVPR 2023)}

This paper introduces a novel approach to generating human motion sequences based on conditional inputs, such as text descriptions or action categories, through a \textbf{Motion Latent Diffusion (MLD)} model \cite{chen2023executing}. The challenge in motion generation arises from the highly diverse nature of human movements and the difficulty of aligning them with conditional inputs like natural language. To address this, a Variational AutoEncoder (VAE) is designed to map human motion sequences into a compact latent space. Instead of applying a diffusion model directly to raw motion sequences—which is computationally expensive and prone to noise—a diffusion process is proposed on the latent motion representations. This reduces both the computational load and the risk of artifacts, while improving the quality of generated motions.

The key advantage of the MLD model is its efficiency, producing high-quality motion sequences with significantly lower computational overhead compared to existing methods. Extensive experiments demonstrate that the MLD model outperforms state-of-the-art techniques across various tasks, such as text-to-motion and action-to-motion generation, achieving faster inference and better diversity in generated motions.

In comparison to traditional approaches, MLD also eliminates the need for large amounts of annotated motion data by leveraging large-scale unannotated motion datasets. The method demonstrates competitive performance across multiple benchmarks, highlighting its potential for diverse applications in gaming, film, robotics, and more.

\begin{figure}[h]
\centering
\includegraphics[width=0.5\textwidth]{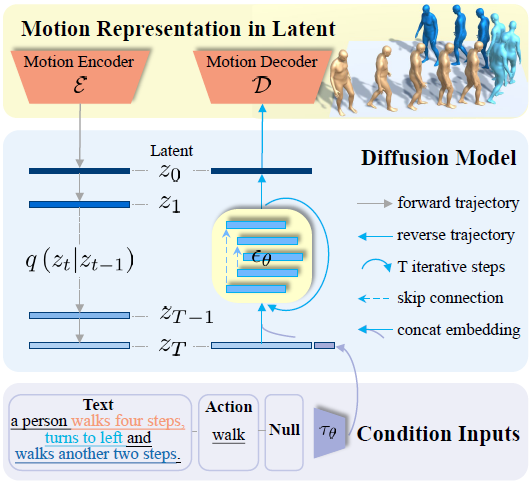} 
\caption{Overview of the Motion Latent Diffusion (MLD) model. The model learns to generate human motion sequences by first encoding motion data into a compact latent space using a Variational AutoEncoder (VAE). The generated latent codes are then processed by a diffusion model conditioned on inputs such as action categories or textual descriptions. The process significantly reduces computational overhead and improves motion diversity.}
\label{fig:mld_model}
\end{figure}

\subsection{T2M-GPT: Generating Human Motion from Textual Descriptions with Discrete Representations (CVPR 2023)}

This paper introduces a simple yet effective framework for generating high-quality human motion from textual descriptions, combining Vector Quantized Variational AutoEncoder (VQ-VAE) and a Generative Pre-trained Transformer (GPT) \cite{zhang2023generating}. The framework, named T2M-GPT, operates in two stages: (1) a motion VQ-VAE encodes 3D human motion into discrete latent representations, and (2) a GPT model generates sequences of discrete tokens conditioned on text embeddings.

To address challenges such as codebook collapse in VQ-VAE, the method employs Exponential Moving Average (EMA) for codebook updates and Code Reset to reactivate unused codes. Additionally, a corruption strategy is applied during GPT training to mitigate discrepancies between training and inference. These strategies enhance training stability and improve the quality of generated motions.

The framework achieves state-of-the-art results on HumanML3D and KIT-ML datasets. On HumanML3D, it achieves an FID of 0.116, significantly outperforming MotionDiffuse (FID: 0.630), while maintaining comparable text-motion consistency (R-Precision). Furthermore, the approach supports diverse and multimodal motion generation, demonstrating its robustness across tasks.

An analysis of dataset size and quantization strategies reveals that larger datasets and improved quantization techniques can further enhance performance. The results highlight the effectiveness of discrete representation learning for text-to-motion tasks, providing a simpler yet competitive alternative to diffusion-based methods.

\begin{figure}[h]
    \centering
    \includegraphics[width=0.8\linewidth]{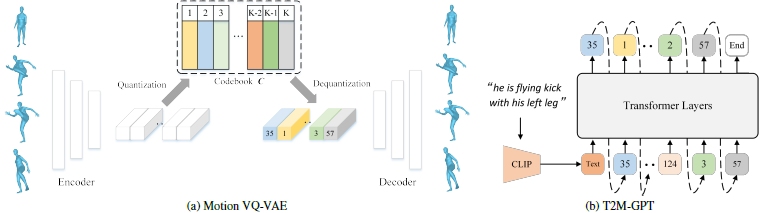}
    \caption{Overview of the T2M-GPT framework. (a) Motion VQ-VAE encodes motion into discrete tokens. (b) T2M-GPT generates token sequences conditioned on text embeddings. The decoder reconstructs motion from tokens.}
    \label{fig:framework}
\end{figure}

\subsection{MMM: Generative Masked Motion Model (CVPR 2024)}

This paper presents MMM, a novel approach for text-driven human motion generation that addresses challenges in speed, fidelity, and editability often encountered in diffusion and autoregressive models \cite{pinyoanuntapong2024mmm}. MMM consists of two main components: a motion tokenizer and a conditional masked motion transformer. The motion tokenizer utilizes a vector quantized variational autoencoder (VQ-VAE) to convert 3D human motion into discrete latent tokens, ensuring fine-grained motion details are preserved. The masked motion transformer predicts masked motion tokens conditioned on text and unmasked motion tokens, enabling parallel and iterative decoding.

By employing bidirectional attention mechanisms, MMM effectively captures semantic correlations between motion and text tokens. This allows for high-quality and efficient motion generation. Additionally, the model supports advanced editing tasks such as motion in-betweening, upper body editing, and long-sequence generation by placing mask tokens in regions requiring modification. Smooth transitions between edited and unedited parts are achieved without additional training.

Experiments on the HumanML3D and KIT-ML datasets demonstrate that MMM achieves state-of-the-art performance, with superior FID scores (0.08 on HumanML3D and 0.429 on KIT-ML) and faster inference speeds compared to existing methods. MMM is two orders of magnitude faster than diffusion-based models and significantly outperforms autoregressive models in both speed and quality. 

\begin{figure}[h]
    \centering
    \includegraphics[width=0.8\linewidth]{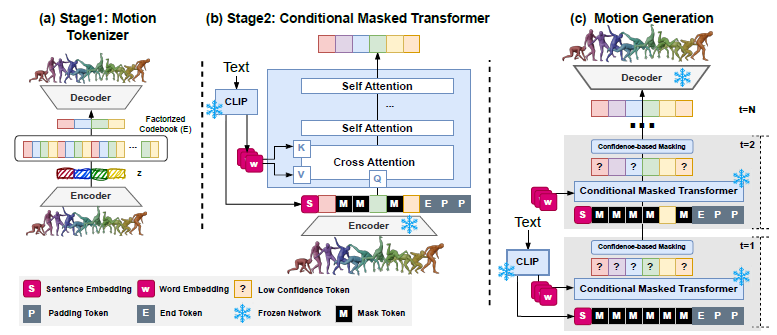} 
    \caption{Overview of MMM. (a) The motion tokenizer transforms raw motion into discrete tokens. (b) The conditional masked transformer predicts masked motion tokens conditioned on text embeddings and unmasked motion tokens. (c) Motion generation is performed iteratively and concurrently.}
    \label{fig:mmm_architecture}
\end{figure}

\subsection{Motion Mamba: Efficient and Long Sequence Motion Generation (ECCV 2024)}

Human motion generation is a crucial task in generative computer vision, yet achieving long-sequence motion with high efficiency remains challenging. Recent advancements in state space models (SSMs), particularly Mamba, have shown promise in long sequence modeling. However, their adaptation to motion generation is hindered by the lack of a specialized architecture for modeling motion sequences.  

To address these challenges, \textbf{Motion Mamba} \cite{zhang2024motion} is introduced as a novel motion generation framework utilizing SSMs \cite{zhang2024kmm,zhang2024infinimotion}. Motion Mamba incorporates:  
(1) A \textbf{Hierarchical Temporal Mamba (HTM)} block that processes temporal motion sequences through a U-Net-based SSM ensemble, ensuring consistent and smooth transitions between frames.  
(2) A \textbf{Bidirectional Spatial Mamba (BSM)} block that models latent spatial dependencies in both forward and backward directions, enhancing pose accuracy and motion fidelity.  

Significant improvements over previous diffusion-based motion generation models are demonstrated. Motion Mamba achieves up to \textbf{50\% improvement in FID} and operates \textbf{four times faster} than the best-performing diffusion model. Evaluations on the HumanML3D and KIT-ML datasets confirm superior generation quality, diversity, and efficiency.  

\begin{figure}[h]
    \centering
    \includegraphics[width=0.8\linewidth]{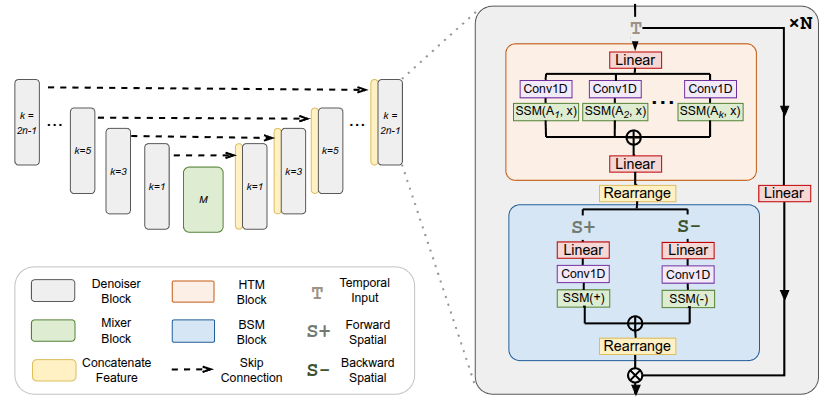}
    \caption{Overview of Motion Mamba architecture. The model integrates Hierarchical Temporal Mamba (HTM) and Bidirectional Spatial Mamba (BSM) blocks within a U-Net-based diffusion framework for efficient long-sequence motion generation.}
    \label{fig:motion_mamba}
\end{figure}

\subsection{EMDM: Efficient Motion Diffusion Model for Fast and High-Quality Motion Generation (ECCV 2024)}

This paper presents the Efficient Motion Diffusion Model (EMDM), which achieves fast and high-quality human motion generation \cite{zhou2024emdm}. Existing motion diffusion models, such as MDM and MotionDiffuse, are hindered by slow inference speeds due to their reliance on thousands of denoising steps. Motion latent diffusion models improve efficiency by operating in a latent space but face challenges in effectively learning the motion embedding space, resulting in quality limitations. To address these issues, EMDM utilizes a conditional denoising diffusion GAN (Generative Adversarial Network) to model complex denoising distributions with fewer sampling steps.

The model incorporates a conditional generator and discriminator, which leverage control signals (e.g., text or action labels) and time-step information to enable efficient motion generation. By capturing multimodal data distributions, EMDM supports larger sampling step sizes, significantly reducing runtime while maintaining high fidelity and diversity in generated motions. Geometric loss functions are further integrated to enhance motion quality and stabilize the training process.

Experiments conducted on HumanML3D, KIT, and HumanAct12 datasets demonstrate that EMDM achieves state-of-the-art performance in text-to-motion and action-to-motion tasks. The model significantly outperforms existing methods in inference speed, achieving a runtime of 0.05 seconds per sequence in text-to-motion tasks, compared to 12.3 seconds for MDM.

\begin{figure}[h]
    \centering
    \includegraphics[width=0.8\linewidth]{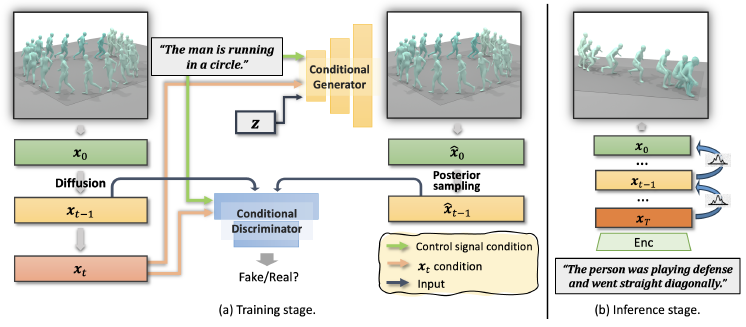} 
    \caption{Pipeline of EMDM. The model employs a conditional denoising diffusion GAN to capture complex denoising distributions, enabling efficient motion sampling with high quality.}
    \label{fig:emdm_pipeline}
\end{figure}

\section{Long Human Motion Generation}

Long Human Motion Generation refers to the task of synthesizing extended sequences of realistic and coherent human motions in a continuous manner. This task is fundamental in computer vision and computer graphics, with applications in animation, virtual reality, robotics, and gaming. Unlike short motion generation, which focuses on producing single or brief actions, long motion generation requires modeling complex temporal dependencies to ensure smooth transitions between actions and maintain realism over extended durations.

The task presents significant challenges due to the high dimensionality of human motion data, the diversity of possible actions, and the need for temporal consistency. It often involves generating a sequence of body poses that evolve naturally over time, guided by input constraints such as text descriptions, action labels, or environmental interactions. Success in this task depends on capturing both the dynamics of individual actions and the transitions between them, ensuring the generated motions align with human behavior.

\subsection{TEACH: Temporal Action Compositions for 3D Humans (3DV 2022)}

This paper presents TEACH (Temporal Action Composition for 3D Humans), a framework designed to generate 3D human motions from sequential natural language descriptions \cite{athanasiou2022teach}. Unlike prior methods that primarily handle single actions or sentences, TEACH addresses the challenge of synthesizing temporally compositional actions, enabling the creation of a sequence of actions with smooth transitions. This capability is crucial for applications such as virtual reality and animation, where realistic and continuous human motion is essential.

The approach utilizes the BABEL dataset, which provides diverse motion sequences paired with free-form language descriptions. TEACH employs a hierarchical architecture, combining autoregressive modeling for sequential action generation with non-autoregressive modeling within individual actions. A past-conditioned text encoder is incorporated to encode the previous motion frames and current textual instructions, ensuring temporal continuity across actions. Additionally, spherical linear interpolation (Slerp) is applied to refine transitions between actions.

Quantitative evaluations on the BABEL dataset demonstrate that TEACH outperforms baseline methods in generating smooth and realistic motion sequences. Metrics such as Average Positional Error (APE) and Average Variational Error (AVE) highlight its effectiveness. Qualitative results further illustrate its ability to handle diverse and complex action sequences, such as "step forward with the right foot" followed by "kick with the left foot." 

\begin{figure}[h]
    \centering
    \includegraphics[width=0.6\linewidth]{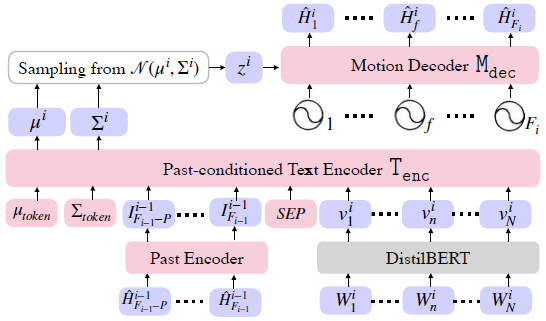} 
    \caption{Overview of the TEACH pipeline. The model integrates a past-conditioned text encoder and a motion decoder to generate sequential 3D human motions from textual descriptions.}
    \label{fig:teach_pipeline}
\end{figure}

\subsection{PriorMDM: Human Motion Diffusion as a Generative Prior (ICLR 2024)}

Recent advancements in diffusion models have demonstrated their potential for human motion generation \cite{shafir2023human}. This paper introduces three innovative motion composition methods: sequential composition, parallel composition, and model composition, leveraging a pretrained Motion Diffusion Model (MDM). These approaches address challenges in generating long sequences, multi-person interactions, and fine-grained motion control, even under limited data scenarios.

1. \textit{Sequential Composition}: The DoubleTake method facilitates the generation of arbitrarily long motion sequences by composing short intervals. It ensures smooth transitions between intervals through a two-stage inference process. The first stage generates overlapping intervals with "handshakes" to maintain continuity, while the second stage refines transitions using a soft-masking mechanism.

2. \textit{Parallel Composition}: For two-person interaction generation, a lightweight communication block, ComMDM, is introduced to coordinate motions between two fixed MDM priors. This approach demonstrates promising results for two-person motion synthesis in a few-shot learning setup.

3. \textit{Model Composition}: DiffusionBlending allows blending multiple fine-tuned models to enable precise control over specific joints or trajectories. This method generalizes classifier-free guidance, enabling cross-model combinations for flexible motion editing.

The proposed methods are evaluated on various tasks, such as long-sequence generation, two-person motion synthesis, and fine-grained motion control. Results indicate that these methods produce high-quality, coherent, and diverse motions, outperforming prior approaches with minimal additional training.

\begin{figure}[h]
    \centering
    \includegraphics[width=0.5\linewidth]{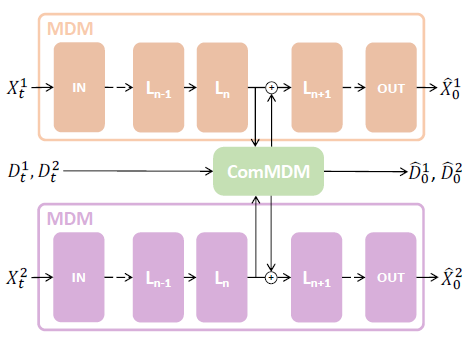} 
    \caption{Overview of the proposed methods: (a) Sequential composition with DoubleTake, (b) Parallel composition with ComMDM, and (c) Model composition with DiffusionBlending.}
    \label{fig:prior_mdm_methods}
\end{figure}

\subsection{MultiAct: long-term 3D human motion generation from multiple action labels (AAAI 2023)}

This paper introduces MultiAct, a novel framework designed to generate long-term 3D human motion from multiple action labels \cite{lee2023multiact}. Existing action-conditioned methods are limited to generating short-term motions for single actions, while motion-conditioned methods generate future motions without user control over desired actions. MultiAct addresses these challenges by combining action and motion conditions in a unified recurrent framework.

1. \textit{Recurrent Framework}: MultiAct generates long-term motions by iteratively taking the previous motion and the next action label as inputs. The framework ensures smooth transitions between actions and produces realistic motions aligned with the input action sequence.

2. \textit{MACVAE Module}: A key component of MultiAct is the Motion- and Action-Conditioned Variational Autoencoder (MACVAE), which concurrently generates action motions and transitions. MACVAE leverages the joint condition of the previous motion and the current action label to ensure transitions are contextually consistent with adjoining motions.

3. \textit{Face-Front Canonicalization}: To address issues with ground geometry during recurrent generation, a face-front canonicalization method is proposed. This technique normalizes input motions while preserving the relative geometry between the body and the ground, ensuring physically plausible outputs.

Experimental results demonstrate that MultiAct significantly outperforms state-of-the-art methods, including combinations of action-conditioned and motion-in-betweening models. MultiAct achieves higher accuracy, diversity, and multimodality in generated motions, while maintaining smooth and realistic transitions.

\begin{figure}[h]
    \centering
    \includegraphics[width=0.8\linewidth]{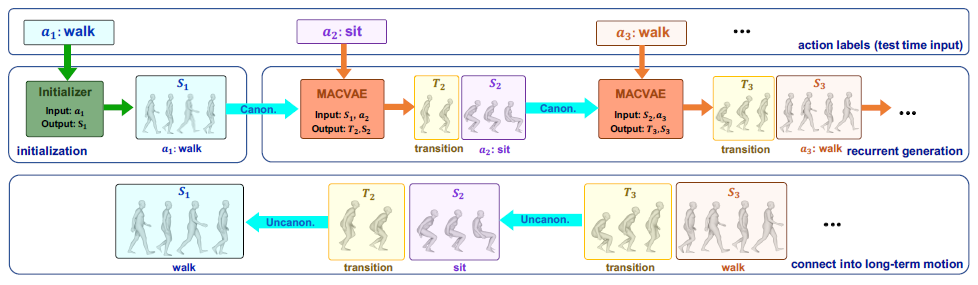}
    \caption{Overview of MultiAct: (a) Input action labels, (b) Recurrent generation with MACVAE, and (c) Long-term motion generation with smooth transitions.}
    \label{fig:multiact_overview}
\end{figure}

\subsection{M2D2M: Multi-Motion Generation from Text with Discrete Diffusion Models (ECCV 2024)}

This paper introduces M2D2M, a novel framework for generating long-term, coherent human motion sequences from textual descriptions of multiple actions \cite{chi2024m2d2m}. Unlike traditional methods that generate motions for single actions and attempt to connect them, M2D2M leverages discrete diffusion models to seamlessly integrate multiple motions while maintaining fidelity to individual action descriptions. A key feature of M2D2M is the dynamic transition probability mechanism, which adapts based on the proximity between motion tokens, ensuring smooth transitions and contextual coherence at motion boundaries. To further enhance motion generation, the framework employs a Two-Phase Sampling (TPS) strategy. In the first phase, joint sampling is used to sketch a coarse outline of the entire sequence, while the second phase refines each motion independently, enabling the generation of smooth multi-motion sequences using models trained only on single-motion data. 

To evaluate the smoothness of transitions, the paper introduces a novel metric, Jerk, which quantifies the smoothness of motion at action boundaries. Experimental results on the HumanML3D and KIT-ML datasets demonstrate that M2D2M achieves state-of-the-art performance, surpassing benchmarks like PriorMDM and T2M-GPT in terms of fidelity, diversity, and transition smoothness. By effectively balancing the fidelity of individual motions and the natural flow of transitions, M2D2M establishes itself as a robust framework for text-driven human motion generation.

\begin{figure}[h]
    \centering
    \includegraphics[width=0.8\linewidth]{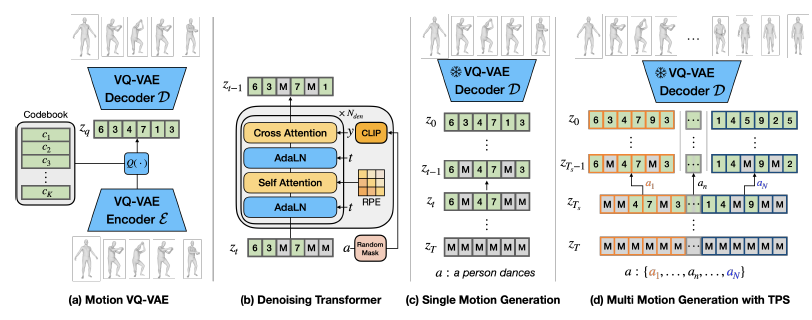} 
    \caption{Overview of M2D2M: Motion tokens are extracted using VQ-VAE, discrete diffusion is applied for denoising, and Two-Phase Sampling (TPS) generates coherent multi-motion sequences.}
    \label{fig:m2d2m_overview}
\end{figure}




\section{Human Hand Object Interaction}

Human Hand Object Interaction (HOI) is a fundamental task in computer vision and computer graphics, focusing on understanding and modeling the interactions between human hands and objects. This task involves capturing and analyzing hand movements, object positions, and the physical contact or dynamic relationships between them. Key aspects of HOI include hand pose estimation, object manipulation trajectory prediction, modeling interaction forces, and analyzing contact points between the hand and objects.

HOI is critical for various applications, such as natural interactions in virtual reality (VR) and augmented reality (AR), robotic grasping and manipulation, animation production, and sign language recognition. However, the task is highly challenging due to the hand's high degrees of freedom, complex geometry, and the diverse shapes, sizes, and materials of objects. Research in HOI not only advances human-computer interaction technologies but also contributes to understanding human behavior.

\subsection{Physical Interaction: Reconstructing Hand-object Interactions
with Physics (SA 2022)}

This paper presents a physics-based method for reconstructing hand-object interactions from a single-view depth camera in real-time \cite{hu2022physical}. The method addresses challenges such as severe occlusions and missing observations, which often lead to ambiguities in traditional kinematic models. By incorporating physics-driven priors, the proposed system achieves physically plausible and accurate reconstructions of hand-object interactions, including contact forces.

The system consists of three stages: kinematic motion tracking, physics-based contact status optimization, and confidence-based contact movement modeling. First, kinematic tracking estimates hand poses and object motions. Then, contact status optimization refines hand-object contacts by leveraging physical rules, such as ensuring object dynamics are consistent with contact forces. Finally, the confidence-based slide prevention module distinguishes between static and sliding contacts by combining kinematic confidence and contact force estimation, effectively handling ambiguous or noisy input data.

Experimental results demonstrate that the method outperforms state-of-the-art techniques in both qualitative and quantitative evaluations. It significantly improves fingertip tracking accuracy, reduces physically implausible artifacts (e.g., insufficient contact points), and achieves real-time performance. The method also estimates plausible contact forces, making it suitable for applications in gaming, virtual reality, human-computer interaction, and robotics.

\begin{figure}[h]
    \centering
    \includegraphics[width=0.8\linewidth]{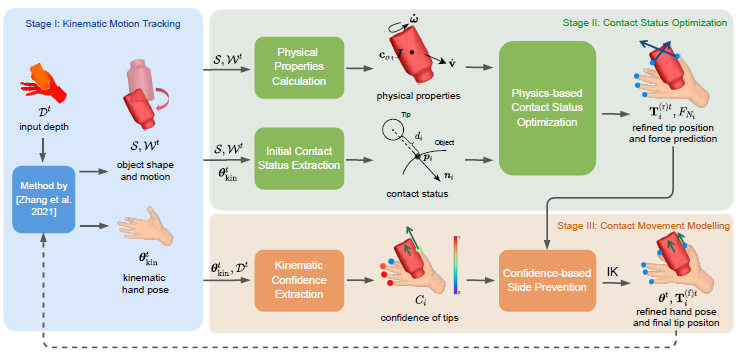} 
    \caption{Overview of the proposed method: (1) Kinematic motion tracking estimates initial hand-object interactions; (2) Physics-based contact status optimization refines contacts using object dynamics; (3) Confidence-based slide prevention models static and sliding contact motions.}
    \label{fig:method_overview}
\end{figure}

\subsection{DeepSimHO: Stable Pose Estimation for Hand-Object Interaction via Physics Simulation (NIPS 2023)}

DeepSimHO is a framework designed to estimate stable and accurate 3D hand-object poses from a single image \cite{wang2024deepsimho}. Unlike prior methods that primarily rely on proximity constraints, this approach emphasizes the dynamic nature of hand-object interaction, ensuring physical stability by counteracting gravity and preventing object slippage.

The framework consists of three main components: a base network for initial hand-object pose estimation, a physics simulator for stability evaluation, and a neural network named DeepSim. The base network generates an initial pose, which is then evaluated by the physics simulator to measure stability based on object displacement under gravity. To address the challenges of unreliable gradients from differentiable simulators, DeepSim approximates the stability evaluation process and provides smooth gradients for backpropagation.

By refining the base network with stability loss predicted by DeepSim, the framework achieves physically plausible and stable results. Experiments on the DexYCB and HO3D datasets demonstrate significant improvements in stability metrics, such as reduced simulation displacement and increased success rate, while maintaining comparable pose accuracy. Furthermore, the framework avoids computationally expensive test-time optimization, making it suitable for real-time applications.

\begin{figure}[h]
    \centering
    \includegraphics[width=0.8\linewidth]{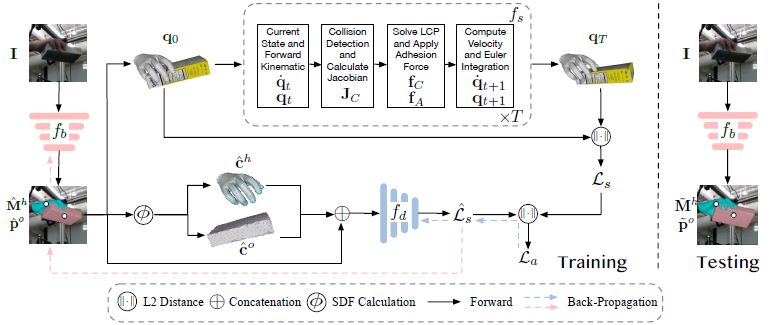} 
    \caption{Overview of DeepSimHO: (1) A base network estimates initial hand-object poses; (2) A physics simulator evaluates stability; (3) DeepSim refines the stability loss to improve pose estimation.}
    \label{fig:deepsimho_overview}
\end{figure}

\subsection{Interacting Hand-Object Pose Estimation via Dense Mutual Attention (WACV 2023)}

This paper introduces a novel framework for 3D hand-object pose estimation from a single monocular image \cite{wang2023interacting}. The proposed approach addresses the challenge of modeling fine-grained hand-object interactions by introducing a dense mutual attention mechanism. Unlike prior methods that rely on sparse keypoint correlations or computationally expensive iterative optimization, this method enables direct feature aggregation between all hand and object vertices, allowing for detailed interaction modeling.

The framework begins by separately estimating rough hand and object meshes using graph-based representations. Hand and object nodes are connected based on their mesh structures, and features are sampled from the input image. The core innovation is the dense mutual attention mechanism, which aggregates features between hand and object graphs. Specifically, each node in the hand graph attends to all nodes in the object graph and vice versa, effectively capturing fine-grained dependencies. This attention mechanism is integrated into a graph convolutional network (GCN) to refine the initial pose estimates.

Quantitative and qualitative experiments on the HO3D and DexYCB datasets demonstrate that the method achieves state-of-the-art performance in terms of accuracy and physical plausibility. The dense mutual attention mechanism outperforms sparse keypoint-based methods by better capturing hand-object interactions, resulting in more realistic and stable pose estimations.

\begin{figure}[h]
    \centering
    \includegraphics[width=0.9\linewidth]{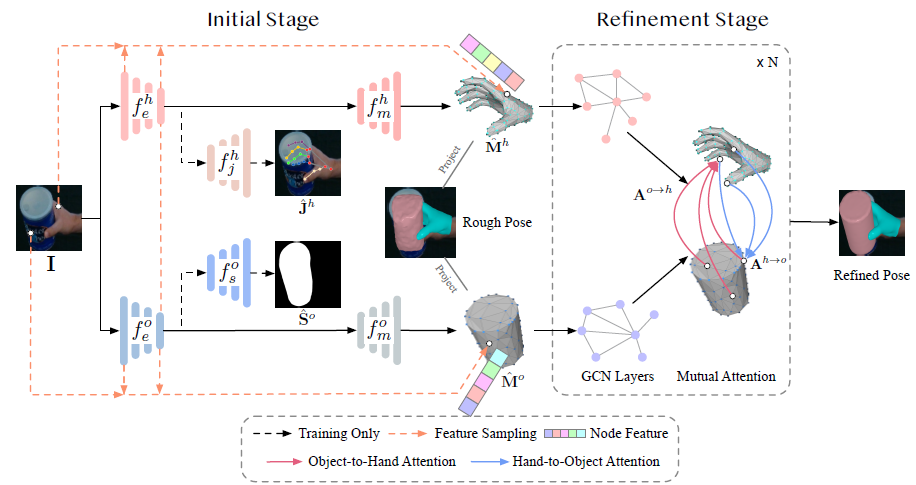}
    \caption{Overview of the proposed framework. Rough hand and object meshes are estimated initially, followed by a refinement stage using graph convolutional layers and the dense mutual attention mechanism to model fine-grained hand-object interactions.}
    \label{fig:dense_mutual_attention}
\end{figure}

\subsection{DiffH2O: Diffusion-Based Synthesis of Hand-Object Interactions from Textual Descriptions (SA 2024)}

DiffH2O introduces a diffusion-based framework to synthesize hand-object interactions from natural language descriptions \cite{christen2024diffh2o}. The method employs a temporal two-stage diffusion process, decoupling the generation into grasping and interaction phases. This design enhances generalization to unseen objects and allows fine-grained control over generated motions. 

The grasping stage models the approach and grasp of a static object, while the interaction stage focuses on manipulating the object based on textual prompts. To ensure smooth transitions between these stages, a novel subsequence imputation technique is proposed. Additionally, grasp guidance is introduced to improve generalization and controllability by directing the diffusion model towards a target grasp.

The framework uses a compact representation that encodes hand-object poses relative to the object’s initial position and includes signed distance fields (SDF) to capture fine details of object geometry. Detailed textual descriptions are also provided for the GRAB dataset, enabling robust text-based control of the generated motions.

Quantitative and qualitative evaluations demonstrate that DiffH2O generates realistic and physically plausible hand-object motions, generalizes well to unseen objects, and significantly outperforms existing methods in terms of diversity, physical accuracy, and semantic alignment with input text prompts.

\begin{figure}[h]
    \centering
    \includegraphics[width=0.9\linewidth]{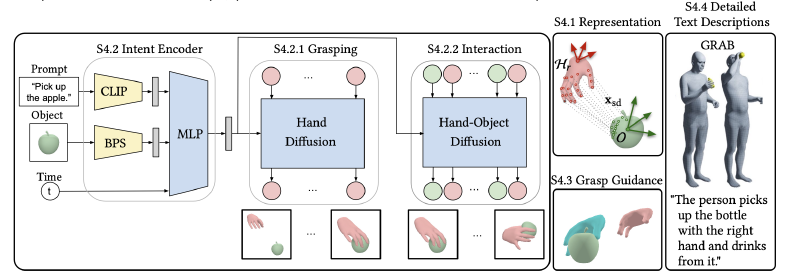}
    \caption{Overview of the DiffH2O framework. The method uses a two-stage diffusion process to generate hand-object interactions. Grasping and interaction stages are connected via subsequence imputation, and the model is guided by textual descriptions and object geometry.}
    \label{fig:diffh2o_overview}
\end{figure}


\section*{Acknowledgments}
We would like to acknowledge the support of Zeyu Zhang and AI Geeks in contributing to this research project.


\begin{thebibliography}{10}

\bibitem{mao2021generating}
Wei Mao, Miaomiao Liu, and Mathieu Salzmann.
\newblock Generating smooth pose sequences for diverse human motion prediction.
\newblock In {\em Proceedings of the IEEE/CVF International Conference on Computer Vision}, pages 13309--13318, 2021.

\bibitem{ionescu2013human3}
Catalin Ionescu, Dragos Papava, Vlad Olaru, and Cristian Sminchisescu.
\newblock Human3. 6m: Large scale datasets and predictive methods for 3d human sensing in natural environments.
\newblock {\em IEEE transactions on pattern analysis and machine intelligence}, 36(7):1325--1339, 2013.

\bibitem{yue2024human}
Jiangbei Yue, Baiyi Li, Julien Pettr{\'e}, Armin Seyfried, and He~Wang.
\newblock Human motion prediction under unexpected perturbation.
\newblock In {\em Proceedings of the IEEE/CVF Conference on Computer Vision and Pattern Recognition}, pages 1501--1511, 2024.

\bibitem{gage2004kinematic}
William~H Gage, David~A Winter, James~S Frank, and Allan~L Adkin.
\newblock Kinematic and kinetic validity of the inverted pendulum model in quiet standing.
\newblock {\em Gait \& posture}, 19(2):124--132, 2004.

\bibitem{louharmonizing}
Zhenyu Lou, Qiongjie Cui, Tuo Wang, Zhenbo Song, Luoming Zhang, Cheng Cheng, Haofan Wang, Xu~Tang, Huaxia Li, and Hong Zhou.
\newblock Harmonizing stochasticity and determinism: Scene-responsive diverse human motion prediction.
\newblock In {\em The Thirty-eighth Annual Conference on Neural Information Processing Systems}.

\bibitem{zheng2022gimo}
Yang Zheng, Yanchao Yang, Kaichun Mo, Jiaman Li, Tao Yu, Yebin Liu, C~Karen Liu, and Leonidas~J Guibas.
\newblock Gimo: Gaze-informed human motion prediction in context.
\newblock In {\em European Conference on Computer Vision}, pages 676--694. Springer, 2022.

\bibitem{araujo2023circle}
Joao~Pedro Ara{\'u}jo, Jiaman Li, Karthik Vetrivel, Rishi Agarwal, Jiajun Wu, Deepak Gopinath, Alexander~William Clegg, and Karen Liu.
\newblock Circle: Capture in rich contextual environments.
\newblock In {\em Proceedings of the IEEE/CVF Conference on Computer Vision and Pattern Recognition}, pages 21211--21221, 2023.

\bibitem{mao2022contact}
Wei Mao, Richard~I Hartley, Mathieu Salzmann, et~al.
\newblock Contact-aware human motion forecasting.
\newblock {\em Advances in Neural Information Processing Systems}, 35:7356--7367, 2022.

\bibitem{goel2023humans}
Shubham Goel, Georgios Pavlakos, Jathushan Rajasegaran, Angjoo Kanazawa, and Jitendra Malik.
\newblock Humans in 4d: Reconstructing and tracking humans with transformers.
\newblock In {\em Proceedings of the IEEE/CVF International Conference on Computer Vision}, pages 14783--14794, 2023.

\bibitem{zhang2024rohm}
Siwei Zhang, Bharat~Lal Bhatnagar, Yuanlu Xu, Alexander Winkler, Petr Kadlecek, Siyu Tang, and Federica Bogo.
\newblock Rohm: Robust human motion reconstruction via diffusion.
\newblock In {\em Proceedings of the IEEE/CVF Conference on Computer Vision and Pattern Recognition}, pages 14606--14617, 2024.

\bibitem{van2024diffusionposer}
Tom Van~Wouwe, Seunghwan Lee, Antoine Falisse, Scott Delp, and C~Karen Liu.
\newblock Diffusionposer: Real-time human motion reconstruction from arbitrary sparse sensors using autoregressive diffusion.
\newblock In {\em Proceedings of the IEEE/CVF Conference on Computer Vision and Pattern Recognition}, pages 2513--2523, 2024.

\bibitem{cong2023efficient}
Lin Cong, Philipp Ruppel, Yizhou Wang, Xiang Pan, Norman Hendrich, and Jianwei Zhang.
\newblock Efficient human motion reconstruction from monocular videos with physical consistency loss.
\newblock In {\em SIGGRAPH Asia 2023 Conference Papers}, pages 1--9, 2023.

\bibitem{guo2020action2motion}
Chuan Guo, Xinxin Zuo, Sen Wang, Shihao Zou, Qingyao Sun, Annan Deng, Minglun Gong, and Li~Cheng.
\newblock Action2motion: Conditioned generation of 3d human motions.
\newblock In {\em Proceedings of the 28th ACM International Conference on Multimedia}, pages 2021--2029, 2020.

\bibitem{lu2022action}
Qiujing Lu, Yipeng Zhang, Mingjian Lu, and Vwani Roychowdhury.
\newblock Action-conditioned on-demand motion generation.
\newblock In {\em Proceedings of the 30th ACM International Conference on Multimedia}, pages 2249--2257, 2022.

\bibitem{xu2023actformer}
Liang Xu, Ziyang Song, Dongliang Wang, Jing Su, Zhicheng Fang, Chenjing Ding, Weihao Gan, Yichao Yan, Xin Jin, Xiaokang Yang, et~al.
\newblock Actformer: A gan-based transformer towards general action-conditioned 3d human motion generation.
\newblock In {\em Proceedings of the IEEE/CVF International Conference on Computer Vision}, pages 2228--2238, 2023.

\bibitem{petrovich2021action}
Mathis Petrovich, Michael~J Black, and G{\"u}l Varol.
\newblock Action-conditioned 3d human motion synthesis with transformer vae.
\newblock In {\em Proceedings of the IEEE/CVF International Conference on Computer Vision}, pages 10985--10995, 2021.

\bibitem{zhang2024motionavatar}
Zeyu Zhang, Yiran Wang, Biao Wu, Shuo Chen, Zhiyuan Zhang, Shiya Huang, Wenbo Zhang, Meng Fang, Ling Chen, and Yang Zhao.
\newblock Motion avatar: Generate human and animal avatars with arbitrary motion.
\newblock {\em arXiv preprint arXiv:2405.11286}, 2024.

\bibitem{chen2023executing}
Xin Chen, Biao Jiang, Wen Liu, Zilong Huang, Bin Fu, Tao Chen, and Gang Yu.
\newblock Executing your commands via motion diffusion in latent space.
\newblock In {\em Proceedings of the IEEE/CVF Conference on Computer Vision and Pattern Recognition}, pages 18000--18010, 2023.

\bibitem{zhang2023generating}
Jianrong Zhang, Yangsong Zhang, Xiaodong Cun, Yong Zhang, Hongwei Zhao, Hongtao Lu, Xi~Shen, and Ying Shan.
\newblock Generating human motion from textual descriptions with discrete representations.
\newblock In {\em Proceedings of the IEEE/CVF conference on computer vision and pattern recognition}, pages 14730--14740, 2023.

\bibitem{pinyoanuntapong2024mmm}
Ekkasit Pinyoanuntapong, Pu~Wang, Minwoo Lee, and Chen Chen.
\newblock Mmm: Generative masked motion model.
\newblock In {\em Proceedings of the IEEE/CVF Conference on Computer Vision and Pattern Recognition}, pages 1546--1555, 2024.

\bibitem{zhang2024motion}
Zeyu Zhang, Akide Liu, Ian Reid, Richard Hartley, Bohan Zhuang, and Hao Tang.
\newblock Motion mamba: Efficient and long sequence motion generation.
\newblock In {\em European Conference on Computer Vision}, pages 265--282. Springer, 2024.

\bibitem{zhang2024kmm}
Zeyu Zhang, Hang Gao, Akide Liu, Qi~Chen, Feng Chen, Yiran Wang, Danning Li, and Hao Tang.
\newblock Kmm: Key frame mask mamba for extended motion generation.
\newblock {\em arXiv preprint arXiv:2411.06481}, 2024.

\bibitem{zhang2024infinimotion}
Zeyu Zhang, Akide Liu, Qi~Chen, Feng Chen, Ian Reid, Richard Hartley, Bohan Zhuang, and Hao Tang.
\newblock Infinimotion: Mamba boosts memory in transformer for arbitrary long motion generation.
\newblock {\em arXiv preprint arXiv:2407.10061}, 2024.

\bibitem{zhou2024emdm}
Wenyang Zhou, Zhiyang Dou, Zeyu Cao, Zhouyingcheng Liao, Jingbo Wang, Wenjia Wang, Yuan Liu, Taku Komura, Wenping Wang, and Lingjie Liu.
\newblock Emdm: Efficient motion diffusion model for fast and high-quality motion generation.
\newblock In {\em European Conference on Computer Vision}, pages 18--38. Springer, 2024.

\bibitem{athanasiou2022teach}
Nikos Athanasiou, Mathis Petrovich, Michael~J Black, and G{\"u}l Varol.
\newblock Teach: Temporal action composition for 3d humans.
\newblock In {\em 2022 International Conference on 3D Vision (3DV)}, pages 414--423. IEEE, 2022.

\bibitem{shafir2023human}
Yonatan Shafir, Guy Tevet, Roy Kapon, and Amit~H Bermano.
\newblock Human motion diffusion as a generative prior.
\newblock {\em arXiv preprint arXiv:2303.01418}, 2023.

\bibitem{lee2023multiact}
Taeryung Lee, Gyeongsik Moon, and Kyoung~Mu Lee.
\newblock Multiact: Long-term 3d human motion generation from multiple action labels.
\newblock In {\em Proceedings of the AAAI Conference on Artificial Intelligence}, volume~37, pages 1231--1239, 2023.

\bibitem{chi2024m2d2m}
Seunggeun Chi, Hyung-gun Chi, Hengbo Ma, Nakul Agarwal, Faizan Siddiqui, Karthik Ramani, and Kwonjoon Lee.
\newblock M2d2m: Multi-motion generation from text with discrete diffusion models.
\newblock In {\em European Conference on Computer Vision}, pages 18--36. Springer, 2024.

\bibitem{hu2022physical}
Haoyu Hu, Xinyu Yi, Hao Zhang, Jun-Hai Yong, and Feng Xu.
\newblock Physical interaction: Reconstructing hand-object interactions with physics.
\newblock In {\em SIGGRAPH Asia 2022 conference papers}, pages 1--9, 2022.

\bibitem{wang2024deepsimho}
Rong Wang, Wei Mao, and Hongdong Li.
\newblock Deepsimho: stable pose estimation for hand-object interaction via physics simulation.
\newblock {\em Advances in Neural Information Processing Systems}, 36, 2024.

\bibitem{wang2023interacting}
Rong Wang, Wei Mao, and Hongdong Li.
\newblock Interacting hand-object pose estimation via dense mutual attention.
\newblock In {\em Proceedings of the IEEE/CVF winter conference on applications of computer vision}, pages 5735--5745, 2023.

\bibitem{christen2024diffh2o}
Sammy Christen, Shreyas Hampali, Fadime Sener, Edoardo Remelli, Tomas Hodan, Eric Sauser, Shugao Ma, and Bugra Tekin.
\newblock Diffh2o: Diffusion-based synthesis of hand-object interactions from textual descriptions.
\newblock In {\em SIGGRAPH Asia 2024 Conference Papers}, pages 1--11, 2024.

\end{thebibliography}

\end{document}